\documentclass[10pt,twocolumn,letterpaper]{article}

\usepackage{cvpr}              

\usepackage{graphicx}
\usepackage{amsmath}
\usepackage{amssymb}
\usepackage{booktabs}
\usepackage{multirow}

\usepackage[pagebackref,breaklinks,colorlinks]{hyperref}

\usepackage[capitalize]{cleveref}
\crefname{section}{Sec.}{Secs.}
\Crefname{section}{Section}{Sections}
\Crefname{table}{Table}{Tables}
\crefname{table}{Tab.}{Tabs.}

\def\cvprPaperID{4317} 
\def\confName{CVPR}
\def\confYear{2022}

\begin{document}

\title{Self-Supervised Predictive Convolutional Attentive Block for Anomaly Detection\vspace*{-0.3cm}}

\author{Nicolae-C\u{a}t\u{a}lin Ristea$^{1,2}$, Neelu Madan$^{3}$, Radu Tudor Ionescu$^{4,5,}$\thanks{corresponding author: raducu.ionescu@gmail.com}\;, Kamal Nasrollahi$^{3,6}$,\\
Fahad Shahbaz Khan$^{2,7}$, Thomas B. Moeslund$^{3}$, Mubarak Shah$^{8}$\\
$^1$University Politehnica of Bucharest, Romania, $^2$MBZ University of Artificial Intelligence, UAE,\\
$^3$Aalborg University, Denmark, $^4$University of Bucharest, Romania, $^5$SecurifAI, Romania,\\
$^6$Milestone Systems, Denmark, $^7$Link\"{o}ping University, Sweden, $^8$University of Central Florida, US\vspace*{-0.3cm}
}
\maketitle

\begin{abstract}
\vspace*{-0.2cm}
Anomaly detection is commonly pursued as a one-class classification problem, where models can only learn from normal training samples, while being evaluated on both normal and abnormal test samples. Among the successful approaches for anomaly detection, a distinguished category of methods relies on predicting masked information (\eg patches, future frames, etc.) and leveraging the reconstruction error with respect to the masked information as an abnormality score. Different from related methods, we propose to integrate the reconstruction-based functionality into a novel self-supervised predictive architectural building block. The proposed self-supervised block is generic and can easily be incorporated into various state-of-the-art anomaly detection methods. Our block starts with a convolutional layer with dilated filters, where the center area of the receptive field is masked. The resulting activation maps are passed through a channel attention module. Our block is equipped with a loss that minimizes the reconstruction error with respect to the masked area in the receptive field. We demonstrate the generality of our block by integrating it into several state-of-the-art frameworks for anomaly detection on image and video, providing empirical evidence that shows considerable performance improvements on MVTec AD, Avenue, and ShanghaiTech. We release our code as open source at: \small{\url{https://github.com/ristea/sspcab}}.
\vspace*{-0.35cm}
\end{abstract}

\section{Introduction}
\label{introduction}

Anomaly detection is an important task with a broad set of applications ranging from industrial inspection (finding defects of objects or materials on industrial production lines) \cite{Bergmann-CVPR-2019,Carrera-TII-2017,Defard-ICPR-2021,Fei-TMM-2020,Li-BMVC-2020,Rudolph-WACV-2021,Salehi-CVPR-2021,Yi-ACCV-2020} to public security (detecting abnormal events such as traffic accidents, fights, explosions, etc.) \cite{Dong-Access-2020,Doshi-CVPRW-2020a,Georgescu-CVPR-2021,Georgescu-TPAMI-2021,Gong-ICCV-2019,Ionescu-WACV-2019,Ji-IJCNN-2020,Lee-TIP-2019,Liu-ICCV-2021,Lu-ECCV-2020,Nguyen-ICCV-2019,Pang-CVPR-2020,Park-CVPR-2020,Ramachandra-WACV-2020a,Ramachandra-PAMI-2020,Sun-ACMMM-2020,Wang-ACMMM-2020,Wu-TNNLS-2019,Yu-ACMMM-2020,Zaheer-CVPR-2020}. The task is typically framed as a one-class classification (outlier detection) problem, where methods \cite{Antic-ICCV-2011,Cheng-CVPR-2015,Dong-Access-2020,Hasan-CVPR-2016,Ionescu-CVPR-2019,Ionescu-WACV-2019,Kim-CVPR-2009,Lee-TIP-2019,Li-PAMI-2014,Liu-CVPR-2018,Lu-ICCV-2013,Luo-ICCV-2017,Mahadevan-CVPR-2010,Mehran-CVPR-2009,Park-CVPR-2020,Ramachandra-WACV-2020a,Ramachandra-WACV-2020b,Ravanbakhsh-WACV-2018,Ravanbakhsh-ICIP-2017,Sabokrou-IP-2017,Tang-PRL-2020,Wu-TNNLS-2019,Xu-CVIU-2017,Zhao-CVPR-2011,Zhang-PR-2020} learn a familiarity model from normal training samples, labeling unfamiliar examples (outliers) as anomalies, at inference time. Since abnormal samples are available only at test time, supervised learning methods are not directly applicable to anomaly detection. To this end, researchers turned their attention to other directions such as reconstruction-based approaches \cite{Fei-TMM-2020,Gong-ICCV-2019,Hasan-CVPR-2016,Li-BMVC-2020,Liu-CVPR-2018,Luo-ICCV-2017,Nguyen-ICCV-2019,Park-CVPR-2020,Ravanbakhsh-ICIP-2017,Salehi-CVPR-2021,Tang-PRL-2020,Venkataramanan-ECCV-2020}, dictionary learning methods \cite{Carrera-TII-2017,Cheng-CVPR-2015,Cong-CVPR-2011, Dutta-AAAI-2015,Lu-ICCV-2013,Ren-BMVC-2015}, distance-based models \cite{Bergmann-CVPR-2020,Defard-ICPR-2021,Ionescu-CVPR-2019,Ionescu-WACV-2019,Ramachandra-WACV-2020a,Ramachandra-WACV-2020b,Ravanbakhsh-WACV-2018,Sabokrou-IP-2017,Sabokrou-CVIU-2018,Saligrama-CVPR-2012,Smeureanu-ICIAP-2017,Sun-PR-2017,Tran-BMVC-2017}, change detection frameworks \cite{Giorno-ECCV-2016,Ionescu-ICCV-2017,Liu-BMVC-2018,Pang-CVPR-2020}, and probabilistic models \cite{Adam-PAMI-2008,Antic-ICCV-2011,Feng-NC-2017,Hinami-ICCV-2017,Kim-CVPR-2009,Mahadevan-CVPR-2010,Mehran-CVPR-2009,Rudolph-WACV-2021,Saleh-CVPR-2013,Wu-CVPR-2010}. 

A distinguished subcategory of reconstruction methods relies on predicting masked information, leveraging the reconstruction error with respect to the masked information as an abnormality score. The masked information can come in different forms, \eg superpixels \cite{Li-BMVC-2020}, future frames \cite{Liu-CVPR-2018}, middle bounding boxes \cite{Georgescu-CVPR-2021}, among others. Methods in this subcategory mask some part of the input and employ a deep neural network to predict the missing input information. Different from such methods, we propose to integrate the capability of reconstructing the masked information into a neural block. Introducing the reconstruction task at a core architectural level has two important advantages: $(i)$ it allows us to mask information at any layer in a neural network (not only at the input), and $(ii)$ it can be integrated into a wide range of neural architectures, thus being very general.

\begin{figure*}[!t]
\begin{center}
\centerline{\includegraphics[width=0.92\linewidth]{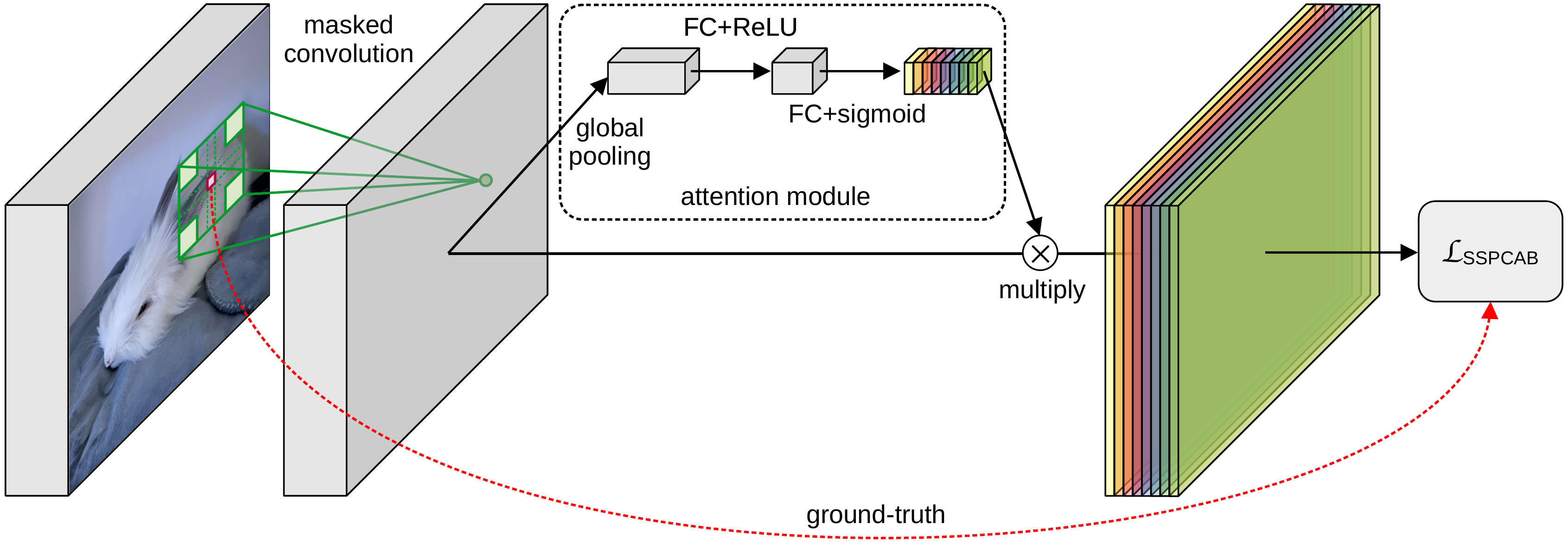}}
\vspace{-0.2cm}
\caption{Our self-supervised predictive convolutional attentive block (SSPCAB). For each location where the dilated convolutional filter is applied, the block learns to reconstruct the masked area using contextual information. A channel attention module performs feature recalibration by using global information to selectively emphasize or suppress reconstruction maps. Best viewed in color.}
\label{fig_pipeline}
\vspace{-0.8cm}
\end{center}
\end{figure*}

We design our reconstruction block as a self-supervised predictive block formed of a dilated convolutional layer and a channel attention mechanism. The dilated filters are based on a custom receptive field, where the center area of the kernel is masked. The resulting convolutional activation maps are then passed through a channel attention module \cite{Hu-CVPR-2018}. The attention module ensures the block does not simply learn to reconstruct the masked region based on linearly interpolating contextual information. Our block is equipped with a loss that minimizes the reconstruction error between the final activation maps and the masked information. In other words, our block is trained to predict the masked information in a self-supervised manner. Our self-supervised predictive convolutional attentive block (SSPCAB) is illustrated in Figure~\ref{fig_pipeline}. For each location where the dilated convolutional filter is applied, the block learns to reconstruct the masked area using contextual information. Meanwhile, the dilation rate becomes a natural way to control the context level (from local to global), as required for the specific application.

We integrate SSPCAB into various state-of-the-art anomaly detection frameworks \cite{Georgescu-TPAMI-2021,Liu-CVPR-2018,Li-CVPR-2021,Liu-ICCV-2021,Park-CVPR-2020,Zavrtanik-ICCV-2021} and conduct comprehensive experiments on the MVTec AD \cite{Bergmann-CVPR-2019}, Avenue \cite{Lu-ICCV-2013} and ShanghaiTech \cite{Luo-ICCV-2017} data sets. Our empirical results show that SSPCAB can bring significant performance improvements, \eg the region-based detection criterion (RBDC) of Liu \etal \cite{Liu-ICCV-2021} on Avenue increases from $41\%$ to $62\%$ by adding SSPCAB. Moreover, with the help of SSPCAB, we are able to report new state-of-the-art performance levels on Avenue and ShanghaiTech. Additionally, we show extra results on the Avenue data set, indicating that the masked convolutional layer can also increase performance levels, all by itself.

In summary, our contribution is twofold:
\begin{itemize}
\vspace{-0.2cm}
    \item We introduce a novel self-supervised predictive convolutional attentive block that is inherently capable of performing anomaly detection. \vspace{-0.2cm}
    \item We integrate the block into several state-of-the-art neural models \cite{Georgescu-TPAMI-2021,Liu-CVPR-2018,Li-CVPR-2021,Liu-ICCV-2021,Park-CVPR-2020,Zavrtanik-ICCV-2021} for anomaly detection, showing significant performance improvements across multiple models and benchmarks.
\end{itemize}

\section{Related Work}
\label{related_work}

As anomalies are difficult to anticipate, methods are typically trained only on normal data, while being tested on both normal and abnormal data \cite{Park-CVPR-2020, Hasan-CVPR-2016}. Therefore, outlier detection \cite{Ionescu-CVPR-2019, Ionescu-WACV-2019,Ramachandra-WACV-2020a,Ramachandra-WACV-2020b,Ravanbakhsh-WACV-2018} and self-supervised learning \cite{Park-CVPR-2020, Gong-ICCV-2019, Liu-ICCV-2021, Georgescu-CVPR-2021, Georgescu-TPAMI-2021, Zavrtanik-ICCV-2021, Lu-ECCV-2020, Li-CVPR-2021} approaches are extensively used to address the anomaly detection task. Anomaly detection methods can be classified into: dictionary learning methods \cite{Carrera-TII-2017,Cheng-CVPR-2015,Cong-CVPR-2011, Dutta-AAAI-2015,Lu-ICCV-2013,Ren-BMVC-2015}, change detection frameworks \cite{Giorno-ECCV-2016,Ionescu-ICCV-2017,Liu-BMVC-2018,Pang-CVPR-2020}, probability-based methods \cite{Adam-PAMI-2008,Antic-ICCV-2011,Feng-NC-2017,Hinami-ICCV-2017,Kim-CVPR-2009,Mahadevan-CVPR-2010,Mehran-CVPR-2009,Rudolph-WACV-2021,Saleh-CVPR-2013,Wu-CVPR-2010}, distance-based models \cite{Bergmann-CVPR-2020,Defard-ICPR-2021,Ionescu-CVPR-2019,Ionescu-WACV-2019,Ramachandra-WACV-2020a,Ramachandra-WACV-2020b,Ravanbakhsh-WACV-2018,Sabokrou-IP-2017,Sabokrou-CVIU-2018,Saligrama-CVPR-2012,Smeureanu-ICIAP-2017,Sun-PR-2017,Tran-BMVC-2017}, and reconstruction-based methods \cite{Fei-TMM-2020,Gong-ICCV-2019,Hasan-CVPR-2016,Li-BMVC-2020,Liu-CVPR-2018,Luo-ICCV-2017,Nguyen-ICCV-2019,Park-CVPR-2020,Ravanbakhsh-ICIP-2017,Salehi-CVPR-2021,Tang-PRL-2020,Venkataramanan-ECCV-2020, Zavrtanik-ICCV-2021}. 

Dictionary-based methods learn the normal behavior by constructing a dictionary, where each entry in the dictionary represents a normal pattern. Ren \etal \cite{Ren-BMVC-2015} extended dictionary learning methods by considering the relation among different entries. Change-detection frameworks detect anomalies by quantifying changes across the video frames, \ie a significant deviation from the immediately preceding event marks the beginning of an abnormal event. After quantifying the change, approaches such as unmasking \cite{Ionescu-ICCV-2017} or ordinal regression \cite{Pang-CVPR-2020} can be used to segregate anomalies. Probability-based methods build upon the assumption that anomalies occur in a low probability region. These methods estimate the probability density function (PDF) of the normal data and evaluate the test samples based on the PDF. For example, Mahadevan \etal \cite{Mahadevan-CVPR-2010} used a Mixture of Dynamic Textures (MDTs) to model the distribution of the spatio-temporal domain, while Rudolph \etal \cite{Rudolph-WACV-2021} employed normalizing flow to represent the normal distribution. Distance-based methods learn a distance function based on the assumption that normal events occur in the close vicinity of the learned feature space, while the abnormal events are far apart from the normal data. For instance, Ramachandra \etal \cite{Ramachandra-WACV-2020b} employed a Siamese network to learn the distance function. Reconstruction-based methods rely on the assumption that the normal examples can be reconstructed more faithfully from the latent manifold. Our new block belongs to the category of reconstruction-based anomaly detection methods, particularly siding with methods that predict or reconstruct missing (or masked) information \cite{Li-BMVC-2020,Liu-CVPR-2018,Georgescu-CVPR-2021}. 

\noindent
\textbf{Reconstruction-based methods.}
In the past few years, reconstruction-based methods became prevalent in anomaly detection. Such methods typically use auto-encoders \cite{Hasan-CVPR-2016} and generative adversarial networks (GANs) \cite{Liu-CVPR-2018}, as these neural models enable the learning of powerful reconstruction manifolds via using normal data only. However, the generalization capability of neural networks sometimes leads to reconstructing abnormal frames with low error \cite{Dong-Access-2020,Georgescu-TPAMI-2021}, affecting the discrimination between abnormal and normal frames. To address this issue, researchers have tried to improve the latent manifold by diversifying the architecture and training methodologies. Some works focusing on transforming the architectures include memory-based auto-encoders \cite{Dong-Access-2020, Park-CVPR-2020, Liu-ICCV-2021}, which memorize the normal prototypes in the training data, thus increasing the discrimination between normal and abnormal samples. Other works remodeled the reconstruction manifold via training the models with pseudo-abnormal samples \cite{Georgescu-TPAMI-2021, Zavrtanik-ICCV-2021, Astrid-ICCVW-2021}. The adversarial training proposed in \cite{Georgescu-CVPR-2021} applies gradient ascent for out-of-domain pseudo-abnormal samples and gradient descent for normal data, thus learning a more powerful discriminative manifold for video anomaly detection. Zavrtanik \etal \cite{Zavrtanik-ICCV-2021} created pseudo-abnormal samples by adding random noise patches on normal images for image anomaly detection. Some variants of auto-encoders, such as Variational Auto-Encoders (VAEs), have been proposed in \cite{Liu-ICCV-2021,Zimmerer-Arxiv-2018} for the anomaly detection task. These works are based on the assumption that VAEs can only reconstruct the normal images. Liu \etal \cite{Liu-ICCV-2021} used a conditional VAE, conditioning the image prediction on optical flow reconstruction, thus accumulating the error from the optical flow reconstruction task with the image prediction. However, this approach can only be applied to video anomaly detection, due to the presence of motion information in the form of optical flow.

\noindent
\textbf{Reconstruction of masked information.} 
A surrogate task for many anomaly detection approaches \cite{Haselmann-ICMLA-2018, Fei-TMM-2020, Liu-CVPR-2018, Yu-ACMMM-2020, Luo-Arxiv-2020} is to erase some information from the input, while making neural networks predict the erased information. Haselmann \etal \cite{Haselmann-ICMLA-2018} framed anomaly detection as an inpainting problem, where patches from images are masked randomly, using the pixel-wise reconstruction error of the masked patches for surface anomaly detection. Fei \etal \cite{Fei-TMM-2020} proposed the Attribute Restoration Network (ARNet), which includes an attribute erasing module (AEM) to disorient the model by erasing certain attributes from an image, such as color and orientation. In turn, ARNet learns to restore the original image and detect anomalies based on the assumption that normal images can be restored properly. The Cloze task \cite{Luo-Arxiv-2020} is about learning to complete a video when certain frames are removed, being recently employed by Yu \etal \cite{Yu-ACMMM-2020} for anomaly detection. In a similar direction, Georgescu \etal \cite{Georgescu-CVPR-2021} proposed middle frame masking as one of the auxiliary tasks for video anomaly detection. Both approaches are based on the assumption that an erased frame can be reconstructed more accurately for regular motion. 
Future frame prediction \cite{Li-CVPR-2021} utilizes past frames to predict the next frame in the video. The anomaly, in this case, is detected through the prediction error. Another approach based on GANs \cite{Sabokrou-ACCV-2018} learns to erase patches from an image, while the discriminator identifies if patches are normal or irregular.

Unlike existing approaches, we are the first to introduce the reconstruction-based functionality as a basic building block for neural architectures. More specifically, we design a novel block based on masked convolution and channel attention to reconstruct a masked part of the convolutional receptive field. As shown in the experiments, our block can be integrated into a multitude of existing anomaly detection frameworks \cite{Georgescu-TPAMI-2021,Liu-CVPR-2018,Li-CVPR-2021,Liu-ICCV-2021,Park-CVPR-2020,Zavrtanik-ICCV-2021}, almost always bringing significant performance improvements.

\section{Method}
\label{method}
 
Convolutional neural networks (CNNs) \cite{lecun-bottou-ieee-1998,Hinton-NIPS-2012} are widely used across a broad spectrum of computer vision tasks, also being prevalent in anomaly detection \cite{Georgescu-TPAMI-2021,Guo-Arxiv-2021,Li-CVPR-2021,Liu-ICCV-2021,Park-CVPR-2020}. CNNs are formed of convolutional layers equipped with kernels which learn to activate on discriminative local patterns, in order to solve a desired task. The local features extracted by a convolutional layer are combined into more complex features by the subsequent convolutional layers. From this learning process, a hierarchy of features emerges, ranging from low-level features (corners, edges, etc.) to high-level features (car wheels, bird heads, etc.) \cite{Zeiler-ECCV-2014}. While this hierarchy of features is extremely powerful, CNNs lack the ability to comprehend the global arrangement of such local features, as noted by Sabour \etal \cite{Sabour-NIPS-2017}. 

In this paper, we introduce a novel self-supervised predictive convolutional attentive block (SSPCAB) that is purposed at learning to predict (or reconstruct) masked information using contextual information. To achieve highly accurate reconstruction results, our block is forced to learn the global structure of the discovered local patterns. Thus, it addresses the issue pointed out in \cite{Sabour-NIPS-2017}, namely the fact that CNNs do not grasp the global arrangement of local features, as they do not generalize to novel viewpoints or affine transformations. To implement this behavior, we design our block as a convolutional layer with dilated masked filters, followed by a channel attention module. The block is equipped with its own loss function, which is aimed at minimizing the reconstruction error between the masked input and the predicted output. 

We underline that our design is generic, as SSPCAB can be integrated into just about any CNN architecture, being able to learn to reconstruct masked information, while offering useful features for subsequent neural layers. Although the capability of learning and using global structure might make SSPCAB useful for a wide range of tasks, we conjecture that our block has a natural and direct applicability in anomaly detection, as explained next. When integrated into a CNN trained on normal training data, SSPCAB will learn the global structure of normal examples only. When presented with an abnormal data sample at inference time, our block will likely provide a poor reconstruction. We can thus measure the quality of the reconstruction and employ the result as a way to differentiate between normal and abnormal examples. In Section~\ref{experiments_and_results}, we provide empirical evidence to support our claims.

SSPCAB is composed of a masked convolutional layer activated by Rectified Linear Units (ReLU) \cite{Nair-ICML-2010}, followed by a Squeeze-and-Excitation (SE) module \cite{Hu-CVPR-2018}. We next present its components in more details.

\begin{figure}[!t]
\begin{center}
\centerline{\includegraphics[width=0.48\linewidth]{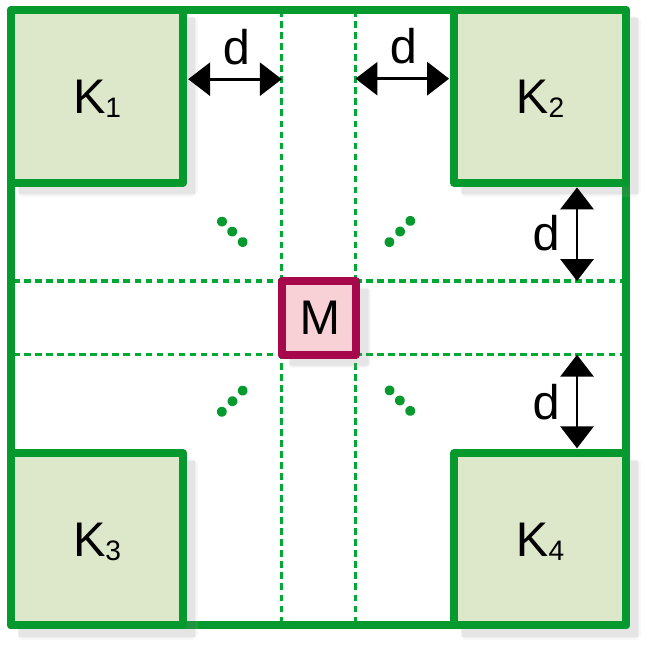}}
\vspace{-0.25cm}
\caption{Our masked convolutional kernel. The visible area of the receptive field is denoted by the regions $\boldsymbol{K}_i, \forall i \in \{1, 2, 3, 4\}$, while the masked area is denoted by $\boldsymbol{M}$. A dilation factor $d$ controls the local or global nature of the visible information with respect to $\boldsymbol{M}$. Best viewed in color.}
\label{fig_kernel}
\vspace{-0.9cm}
\end{center}
\end{figure}

\noindent
{\bf Masked convolution.}
The receptive field of our convolutional filter is depicted in Figure~\ref{fig_kernel}. The learnable parameters of our masked convolution are located in the corners of the receptive field, being denoted by the sub-kernels $\boldsymbol{K}_i \in \mathbb{R}^{k' \times k' \times c}$, $\forall i \in \{1, 2, 3, 4\}$, where $k' \in \mathbb{N}^+$ is a hyperparameter defining the sub-kernel size and $c$ is the number of input channels. Each kernel $\boldsymbol{K}_i$ is located at a distance (dilation rate) $d \in \mathbb{N}^+$ from the masked region in the center of our receptive field, which is denoted by $\boldsymbol{M} \in \mathbb{R}^{1 \times 1 \times c}$. Consequently, the spatial size $k$ of our receptive field can be computed as follows: $k = 2k' + 2d + 1$.

Let $\boldsymbol{X} \in \mathbb{R}^{h \times w \times c}$ be the input tensor of our masked convolutional layer, where $c$ is the number of channels, and $h$ and $w$ are the height and width, respectively.
The convolutional operation performed with our custom kernel in a certain location of the input $\boldsymbol{X}$ only considers the input values from the positions where the sub-kernels $\boldsymbol{K}_i$ are located, the other information being ignored.
The results of the convolution operations between each $\boldsymbol{K}_i$ and the corresponding inputs are summed into a single number, as if the sub-kernels $\boldsymbol{K}_i$ belong to a single convolutional kernel. The resulting value denotes a prediction located at the same position as $\boldsymbol{M}$. Naturally, applying the convolution with one filter produces a single activation map. Hence, we would only be able to predict one value from the masked vector $\boldsymbol{M}$, at the current location. To predict a value for every channel in $\boldsymbol{M}$, we introduce a number of $c$ masked convolutional filters, each predicting the masked information from a distinct channel. As we aim to learn and predict the reconstruction for every spatial location of the input, we add zero-padding of $k'+d$ pixels around the input and set the stride to $1$, such that every pixel in the input is used as masked information. Therefore, the spatial dimension of the output tensor $\boldsymbol{Z}$ is identical to that of the input tensor $\boldsymbol{X}$. Finally, the output tensor is passed through a ReLU activation. We underline that the only configurable hyperparameters of our custom convolutional layer are $k'$ and $d$.


\noindent
{\bf Channel attention module.}
Next, the output of the masked convolution is processed by a channel attention module, which computes an attention score for each channel. Knowing that each activation map in $\boldsymbol{Z}$ is predicted by a separate filter in the presence of masked information, we infer that the masked convolution might end up producing activation maps containing disproportionate (uncalibrated) values across channels. Therefore, we aim to exploit the relationships between channels, with the goal of scaling each channel in $\boldsymbol{Z}$ in accordance with the quality of the representations produced by the masked convolutional layer. To this end, we employ the channel attention module proposed by Hu \etal \cite{Hu-CVPR-2018}. The SE module \cite{Hu-CVPR-2018} provides a mechanism that performs adaptive recalibration of channel-wise feature responses. Through this mechanism, it can learn to use global information to selectively emphasize or suppress reconstruction maps, as necessary. Another motivation to use attention is to increase the modeling capacity of SSPCAB and enable a non-linear processing between the input and output of our block.

Formally, the channel attention block reduces $\boldsymbol{Z}$ to a vector $\boldsymbol{z} \in \mathbb{R}^{c}$ through a global pooling performed on each channel. Subsequently, the vector of scale factors $\boldsymbol{s} \in \mathbb{R}^{c}$ is computed as follows:
\begin{equation}
    \boldsymbol{s} = \sigma\left(\boldsymbol{W}_2 \cdot \delta\left(\boldsymbol{W}_1 \cdot \boldsymbol{z}\right)\right) ,
\end{equation}
where $\sigma$ is the sigmoid activation, $\delta$ is the ReLU activation, and $\boldsymbol{W}_1 \in \mathbb{R}^{\frac{c}{r} \times c}$ and $\boldsymbol{W}_2 \in \mathbb{R}^{c \times \frac{c}{r}}$ represent the weight matrices of two consecutive fully connected (FC) layers, respectively. The first FC layer consists of $\frac{c}{r}$ neurons, squeezing the information by a reduction ratio of $r$. 

Next, the vector $\boldsymbol{s}$ is replicated in the spatial dimension, generating a tensor $\boldsymbol{S}$ of the same size as $\boldsymbol{Z}$. Our last step is the element-wise multiplication between $\boldsymbol{S}$ and $\boldsymbol{Z}$, producing the final tensor $\boldsymbol{\hat{X}} \in \mathbb{R}^{h \times w \times c}$ containing recalibrated features maps.


\noindent
{\bf Reconstruction loss.}
We add a self-supervised task consisting of reconstructing the masked region inside our convolutional receptive field, for every location where the masked filters are applied. To this end, our block should learn to provide the corresponding reconstructions as the output $\boldsymbol{\hat{X}}$. Let $G$ denote the SSPCAB function. We define the self-supervised reconstruction loss as the mean squared error (MSE) between the input and the output, as follows:
\begin{equation}
    \mathcal{L}_{\mbox{\scriptsize{SSPCAB}}}(G, \boldsymbol{X}) =  \left(G(\boldsymbol{X}) - \boldsymbol{X}\right)^2 = \left(\boldsymbol{\hat{X}} - \boldsymbol{X}\right)^2.
\end{equation}

When integrating SSPCAB into a neural model $F$ having its own loss function $\mathcal{L}_F$, our loss can simply be added to the respective loss, resulting in a new loss function that comprises both terms:
\begin{equation}\label{eq_loss_total}
\mathcal{L}_{\mbox{\scriptsize{total}}} = \mathcal{L}_F + \lambda \cdot \mathcal{L}_{\mbox{\scriptsize{SSPCAB}}},
\end{equation}
where $\lambda \in \mathbb{R}^+$ is a hyperparameter that controls the importance of our loss with respect to $\mathcal{L}_F$. We adopt this procedure when incorporating SSPCAB into various neural architectures during our experiments.

\section{Experiments and Results}
\label{experiments_and_results}

\subsection{Data Sets}

\noindent
\textbf{MVTec AD.} The MVTec AD \cite{Bergmann-CVPR-2019} data set is a standard benchmark for evaluating anomaly detection methods on industrial inspection images. It contains images from 10 object categories and 5 texture categories, having 15 categories in total. 
There are 3629 defect-free training images and 1725 test images with or without anomalies.

\noindent
\textbf{Avenue.} The CHUK Avenue \cite{Lu-ICCV-2013} data set is a popular benchmark for video anomaly detection. It contains 16 training and 21 test videos. The anomalies are present only at inference time and include people throwing papers, running, dancing, loitering, and walking in the wrong direction.

\noindent
\textbf{ShanghaiTech.} The ShanghaiTech \cite{Luo-ICCV-2017} benchmark is one of the largest data sets for video anomaly detection. It is formed of 330 training and 107 test videos. As for Avenue, the training videos contain only normal samples, but the test videos can contain both normal and abnormal events. Some examples of anomalies are: people fighting, stealing, chasing, jumping, and riding bike or skating in pedestrian zones.

\subsection{Evaluation Metrics}

\noindent
\textbf{Image anomaly detection.} On MVTec AD, we evaluate methods in terms of the average precision (AP) and the area under the receiver operating characteristic curve (AUROC). The ROC curve is obtained by plotting the true positive rate (TPR) versus the false positive rate (FPR). We consider both localization and detection performance rates. For the detection task, the TPR and FPR values are computed at the image level, \ie TPR is the percentage of anomalous images that are correctly classified, while FPR is the percentage of normal images mistakenly classified as anomalous. For the localization (segmentation) task, TPR is the percentage of abnormal pixels that are correctly classified, whereas FPR is the percentage of normal pixels wrongly classified as anomalous. To determine the segmentation threshold for each method, we follow the approach described in \cite{Bergmann-CVPR-2019}. 

\noindent
\textbf{Video anomaly detection.} We evaluate abnormal event detection methods in terms of the area under the curve (AUC), which is computed by marking a frame as abnormal if at least one pixel inside the frame is abnormal. Following \cite{Georgescu-TPAMI-2021}, we report both the macro and micro AUC scores. The micro AUC is computed after concatenating all frames from the entire test set, while the macro AUC is the average of the AUC scores on individual videos. The frame-level AUC can be an unreliable evaluation measure, as it may fail to evaluate the localization of anomalies \cite{Ramachandra-WACV-2020a}. Therefore, we also evaluate models in terms of the region-based detection criterion (RBDC) and track-based detection criterion (TBDC), as proposed by Ramachandra \etal \cite{Ramachandra-WACV-2020a}. RBDC takes each detected region into consideration, marking a detected region as \emph{true positive} if the Intersection-over-Union with the ground-truth region is greater than a threshold $\alpha$. TBDC measures whether abnormal regions are accurately tracked across time. It considers a detected track as \emph{true positive} if the number of detections in a track is greater than a threshold $\beta$. Following \cite{Ramachandra-WACV-2020a, Georgescu-TPAMI-2021}, we set $\alpha=0.1$ and $\beta=0.1$.


\subsection{Implementation Choices and Tuning}

For the methods \cite{Georgescu-TPAMI-2021,Liu-CVPR-2018,Li-CVPR-2021,Liu-ICCV-2021,Park-CVPR-2020,Zavrtanik-ICCV-2021} chosen to serve as underlying models for SSPCAB, we use the official code from the repositories provided by the corresponding authors, inheriting the hyperparameters, \eg the number of epochs and learning rate, from each method. Unless specified otherwise, we replace the penultimate convolutional layer with SSPCAB in all underlying models.

In a set of preliminary trials with a basic auto-encoder on Avenue, we tuned the hyperparameter $\lambda$ from Eq.~\eqref{eq_loss_total}, representing the weight of the SSPCAB reconstruction error, considering values between $0.1$ and $1$, at a step of $0.1$. Based on these preliminary trials, we decided to use $\lambda=0.1$ across all models and data sets. However, we observed that $\lambda=0.1$ gives a higher than necessary magnitude to our loss for the framework of Liu \etal \cite{Liu-ICCV-2021}. Hence, for Liu \etal \cite{Liu-ICCV-2021}, we reduced $\lambda$ to $0.01$.

\subsection{Preliminary Results}


\begin{table}[t!]
\centering 
\setlength\tabcolsep{3.0pt}
\small
\begin{tabular}{| c | c | c | c | c | c | c | c | c | c |} 
\hline

 \multirow{3}{*}[-0.2ex]{\rotatebox{90}{Method}} & \multirow{2}{*}[0.5ex]{Loss} & \multirow{3}{*}[0.5ex]{$d$} & \multirow{3}{*}[0.5ex]{$k'$} & \multirow{3}{*}[0.5ex]{$r$} & \multirow{2}{*}[0.5ex]{Attention} & \multicolumn{2}{|c|}{AUC} & \multirow{3}{*}[0.5ex]{RBDC} & \multirow{3}{*}[0.5ex]{TBDC} \\
 \cline{7-8}
 & \multirow{2}{*}[0.5ex]{type} &  &  & & \multirow{2}{*}[0.5ex]{type} & \multirow{2}{*}[0.5ex]{Micro} & \multirow{2}{*}[0.5ex]{Macro} &  &  \\
 &  &  &  &  & &  &  &  &  \\
 \hline
 \hline
 \multirow{25}{*}[8.5ex]{\rotatebox{90}{Plain auto-encoder}} & -  & -  & -  & -  & - & 80.0 & 83.4 & 49.98 & 51.69 \\
  \cline {2-10}
  & \multirow{3}{*}[0.0ex]{MAE}  & 0 & 1 & - & - & 83.3 & 84.1 & 47.46 & 52.11 \\
  &  & 1 & 1 & - & - & 83.9 & 84.6 & 49.05 & 52.21 \\
  &  & 2 & 1 & - & - & 83.2 & 84.3 & 48.56 & 52.03 \\
   \cline {2-10}
  & \multirow{3}{*}[0.0ex]{MSE} & 0 & 1 & - & - & 83.6 & 84.2 & 47.86 & 52.21 \\
  &  & 1 & 1 & - & - & 84.2 & 84.9 & 49.22 & 52.29 \\
  &  & 2 & 1 & - & - & 83.6 & 84.3 & 48.44 & 51.98 \\
  \cline {2-10}
  & \multirow{3}{*}[0.0ex]{MSE} & 0 & 2 & - & - & 83.7 & 84.0 & 47.41 & 53.02 \\
  &  & 1 & 2 & - & - & 84.0 & 85.1 & 48.22 & 51.84 \\
  &  & 2 & 2 & - & - & 82.7 & 83.1 & 46.94 & 50.22 \\
  \cline {2-10}
  & \multirow{3}{*}[0.0ex]{MSE} & 0 & 3 & - & - & 82.6 & 83.7 & 48.28 & 51.91 \\
  &  & 1 & 3 & - & - & 82.9 & 84.7 & 48.13 & 52.07 \\
  &  & 2 & 3 & - & - & 83.1 & 83.8 & 47.13 & 49.96 \\
  \cline {2-10}
  & \multirow{3}{*}[0.0ex]{MSE} & 1 & 1 & 8 & CA & \textbf{85.9} & \textbf{85.6} & 53.81 & \textbf{56.33} \\
  &  & 1 & 1 & - & SA & 84.3 & 84.4 & 53.31 & 53.41 \\
  &  & 1 & 1 & 8 & CA+SA & 85.7 & 85.6 & \textbf{53.98} & 54.11 \\
 \cline {2-10}
  & \multirow{2}{*}[0.0ex]{MSE} & 1 & 1 & 4 & CA & 85.6 & 85.3 & 53.83 & 55.99 \\
  &  & 1 & 1 & 16 & CA & 84.4 & 84.9 & 53.28 & 54.37 \\
 \hline
\end{tabular}
\vspace{-0.2cm}
\caption{Micro AUC, macro AUC, RBDC and TBDC scores (in \%) obtained on the Avenue data set with different hyperparameter configurations, \ie kernel size ($k'$), dilation rate ($d$), reduction ratio ($r$), loss type, and attention type, for our SSPCAB. Results are obtained by introducing SSPCAB into a plain auto-encoder that follows the basic architecture designed by Georgescu \etal \cite{Georgescu-TPAMI-2021}. Best results are highlighted in bold.}
\label{tab_ablation}
\vspace{-0.4cm}
\end{table}

We performed preliminary experiments on Avenue to decide the hyperparameters of our masked convolution, \ie the kernel size $k'$ and dilation rate $d$. We consider values in $\{1,2,3\}$ for $k'$, and values in $\{0,1,2\}$ for $d$. 
In addition, we consider two alternative loss functions, namely the Mean Absolute Error (MAE) and Mean Squared Error (MSE), and several types of attention to be added after the masked convolution, namely channel attention (CA), spatial attention (SA), and both (CA+SA).

For the preliminary experiments, we take the appearance convolutional auto-encoder from \cite{Georgescu-TPAMI-2021} as our baseline, stripping out the additional components such as optical flow, skip connections, adversarial training, mask reconstruction and binary classifiers. Our aim is to test various SSPCAB configurations on top of a basic architecture, without trying to overfit the configuration to a specific framework, such as that of Georgescu \etal \cite{Georgescu-TPAMI-2021}. To this end, we decided to remove the aforementioned components, thus using only a plain auto-encoder in our preliminary experiments.

The preliminary results are presented in Table \ref{tab_ablation}. Upon adding the masked convolutional layer based on the MAE loss on top of the basic architecture, we observe significant performance gains, especially for $k'=1$ and $d=1$. The performance further increases when we replace the MAE loss function with MSE. We performed extensive experiments with different combinations of $k'$ and $d$, obtaining better results with $k'=1$ and $d=1$. We therefore decided to fix the loss to MSE, the sub-kernel size $k'$ to $1$, and the dilation rate $d$ to $1$, for all subsequent experiments.
Next, we introduced various attention modules after our masked convolution. Among the considered attention modules, we observe that channel attention is the one that better compliments our masked convolutional layer, providing the highest performance gains for three of the metrics: $5.9\%$ for the micro AUC, $2.2\%$ for the macro AUC, and $4.6\%$ for TBDC. Accordingly, we selected the channel attention module for the remaining experiments. Upon choosing to use channel attention, we test additional reduction rates ($r=4$ and $r=16$), without observing any improvements. As such, we keep the reduction rate of the SE module to $r=8$, whenever we integrate SSPCAB into a neural model.


\begin{figure}[!t]
\begin{center}
\centerline{\includegraphics[width=1.0\linewidth]{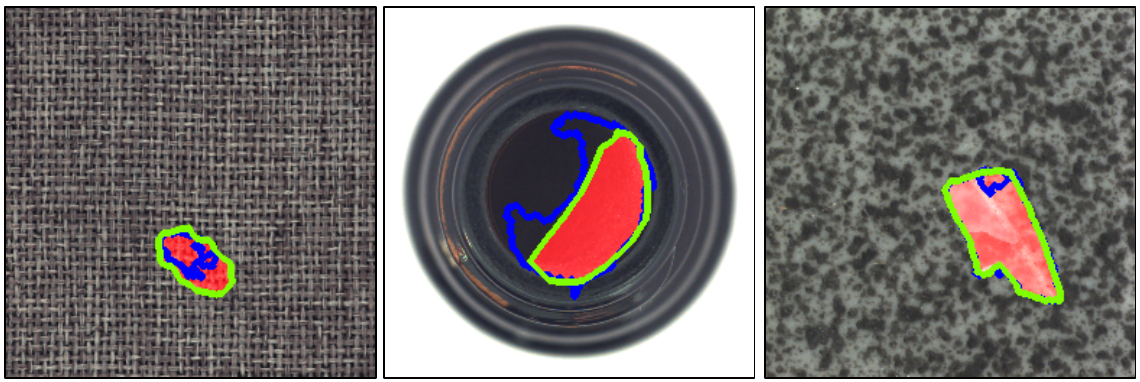}}
\vspace{-0.2cm}
\caption{Anomaly localization examples of DRAEM \cite{Zavrtanik-ICCV-2021} (blue) versus DRAEM+SSPCAB (green) on MVTec AD. The ground-truth anomalies are marked with a red mask. Best viewed in color.}
\label{fig_results_mvtec}
\vspace{-0.8cm}
\end{center}
\end{figure}

\begin{table*}[ht!]
\centering 
\setlength\tabcolsep{4.0pt}
\small
\begin{tabular}{| c | l | cccc | cc | cc | cc | } 
\hline

 & \multirow{5}{*}{Class} & 
\multicolumn{4}{c|}{Localization} & 
\multicolumn{6}{c|}{Detection}  \\ 
\cline{3-12}
 & & \multicolumn{4}{c|}{\multirow{2}{*}{DRAEM \cite{Zavrtanik-ICCV-2021}}}  & \multicolumn{2}{c|}{\multirow{2}{*}{DRAEM \cite{Zavrtanik-ICCV-2021}}}  & \multicolumn{4}{c|}{CutPaste \cite{Li-CVPR-2021}}  \\
  \cline{9-12}
 & & & &  & &  & & \multicolumn{2}{c|}{3-way}  & \multicolumn{2}{c|}{Ensemble} \\ 
  \cline{3-12}
 &  &  & \begin{tabular}[c]{@{}l@{}}+SSPCAB \end{tabular} &  & \begin{tabular}[c]{@{}l@{}}+SSPCAB \end{tabular} &  & \begin{tabular}[c]{@{}l@{}} +SSPCAB \end{tabular} &  & \begin{tabular}[c]{@{}l@{}}+SSPCAB \end{tabular} &  & \begin{tabular}[c]{@{}l@{}}+SSPCAB \end{tabular} \\
   \cline{3-12}
 &  & AUROC  & AUROC  & AP  & AP  & AUROC  & AUROC & AUROC & AUROC & AUROC & AUROC \\
\hline
\hline
\multirow{5}{*}[0.0ex]{\rotatebox{90}{Texture}} & Carpet  &   \textbf{95.5} & 95.0 & 53.5 & \textbf{59.4} &  97.0  &  \textbf{98.2} &   \textbf{93.1}    &  90.7     &   93.9      &   \textbf{96.8}\\

& Grid       & \textbf{99.7} &  99.5 & \textbf{65.7} & 61.1 & 99.9 & \textbf{100.0}   & 99.9      &  99.9     &   \textbf{100.0}        & 99.9\\

& Leather     &  98.6 & \textbf{99.5} & 75.3 & \textbf{76.0} &  100.0  &  100.0 &  100.0    &  100.0     &   100.0      &   100.0\\

& Tile        &  99.2 & \textbf{99.3} & 92.3 & \textbf{95.0} &  99.6  & \textbf{100.0}  & 93.4      &  \textbf{94.0}     &   94.6        & \textbf{95.0}\\

& Wood       & 96.4 & \textbf{96.8} & 77.7 & 77.1 &  99.1  &  \textbf{99.5} & 98.6      &  \textbf{99.2}     &   99.1       & 99.1\\
\hline

\multirow{10}{*}[0.0ex]{\rotatebox{90}{Object}} & Bottle   &   \textbf{99.1}  & 98.8 & 86.5 & \textbf{87.9}  &  \textbf{99.2} &  98.4 &   98.3    &   \textbf{98.6}    &    98.2     &  \textbf{99.1}\\

& Cable      & 94.7 & \textbf{96.0} & 52.4 & \textbf{57.2} &  91.8  & \textbf{96.9}  & 80.6      &  \textbf{82.9}     &   81.2        & \textbf{83.6} \\

& Capsule     & \textbf{94.3} & 93.1 & 49.4  & \textbf{50.2} &  98.5  &  \textbf{99.3} &   96.2    &  \textbf{98.1}    &    \textbf{98.2}      &  97.6\\

& Hazelnut    & 99.7 & \textbf{99.8} & 92.9 & 92.6 & 100.0   & 100.0  & 97.3      &  \textbf{98.3}     &   98.3        & \textbf{98.4}\\

& Metal Nut    & \textbf{99.5} & 98.9 & 96.3 & \textbf{98.1} &  98.7  &  \textbf{100.0} & 99.3      &  \textbf{99.6}     &   99.9        &  99.9\\

& Pill        &  \textbf{97.6} & 97.5 & 48.5 & \textbf{52.4} &  98.9  &  \textbf{99.8} &  92.4     &   \textbf{95.3}    &   94.9        & \textbf{96.6}\\

& Screw      & 97.6 & \textbf{99.8} & 58.2 & \textbf{72.0} &  93.9  &  \textbf{97.9} & 86.3      &  \textbf{90.8}     &   88.7       & \textbf{90.8}\\

& Toothbrush  & 98.1 & 98.1 & 44.7 & \textbf{51.0} &  100.0  & 100.0  & 98.3      &  \textbf{98.8}     &  99.4        & \textbf{99.6}\\

& Transistor   &  \textbf{90.9} & 87.0 & \textbf{50.7} & 48.0 &  \textbf{93.1}  & 92.9  &  95.5     &   \textbf{96.5}    &   96.1        &  \textbf{97.3}\\

& Zipper      & 98.8 & \textbf{99.0} & \textbf{81.5} & 77.1 &  100.0  &  100.0 & \textbf{99.4}      &   99.1    &   99.9        & 99.9\\
\hline
& Overall          & \textbf{97.3} & 97.2 & 68.4 & \textbf{69.9} &  98.0  &  \textbf{98.9} & 95.2      &   \textbf{96.1}   &   96.1        & \textbf{96.9}  \\
\hline
\end{tabular}
\vspace{-0.1cm}
\caption{Localization AUROC/AP and detection AUROC (in \%) of state-of-the-art methods on MVTec AD, before and after adding SSPCAB. The best result for each before-versus-after pair is highlighted in bold.}
\label{table:ImageAnomaly} 
\vspace{-0.3cm}
\end{table*}

\subsection{Anomaly Detection in Images}

\noindent
\textbf{Baselines.}
We choose two recent models for image anomaly detection, \ie \textbf{CutPaste} \cite{Li-CVPR-2021} and \textbf{DRAEM} \cite{Zavrtanik-ICCV-2021}. 

Li \etal \cite{Li-CVPR-2021} proposed \emph{CutPaste}, a simple data augmentation technique that cuts a patch from an image and pastes it to a random location. The CutPaste architecture is built on top of GradCAM \cite{Selvaraju-ICCV-2017}. The model is based on a self-supervised 3-way classification task, learning to classify samples into normal, CutPaste and CutPaste-Scar, where a scar is a long and thin mark of a random color. Li \etal \cite{Li-CVPR-2021} also used an ensemble of five 3-way CutPaste models trained with different random seeds to improve results.

Zavrtanik \etal \cite{Zavrtanik-ICCV-2021} introduced \emph{DRAEM}, a method based on a dual auto-encoder for anomaly detection and localization on MVTec AD. We introduce SSPCAB into both the localization and detection networks.

\noindent
\textbf{Results.} We present the results on MVTec AD in Table~\ref{table:ImageAnomaly}. Considering the detection results, we observe that SSPCAB brings consistent performance improvements on most categories for both CutPaste \cite{Li-CVPR-2021} and DRAEM \cite{Zavrtanik-ICCV-2021}. Moreover, the overall performance gains in terms of detection AUROC are close to $1\%$, regardless of the underlying model. Given that the baselines are already very good, we consider the improvements brought by SSPCAB as noteworthy.

Considering the localization results, it seems that SSPCAB is not able to improve the overall AUROC score of DRAEM \cite{Zavrtanik-ICCV-2021}. However, the more challenging AP metric tells a different story. Indeed, SSPCAB increases the overall AP of DRAEM \cite{Zavrtanik-ICCV-2021} by $1.5\%$, from $68.4\%$ to $69.9\%$.

In Figure~\ref{fig_results_mvtec}, we illustrate a few anomaly localization examples where SSPCAB introduces significant changes to the anomaly localization contours of DRAEM \cite{Zavrtanik-ICCV-2021}, showing a higher overlap with the ground-truth anomalies. We believe that these improvements are a direct effect induced by the reconstruction errors produced by our novel block. We provide more anomaly detection examples in the supplementary.

\begin{table*}[ht!]
\centering 
\setlength\tabcolsep{2.0pt}
\small
\begin{tabular}{| l | l | c | c | c | c | c | c | c | c |} 
\hline
 \multirow{3}{*}{Venue} & \multirow{3}{*}{Method} & \multicolumn{4}{c|}{Avenue} & \multicolumn{4}{c|}{ShanghaiTech}\\
 \cline{3-10}
& & \multicolumn{2}{c|}{AUC} & RBDC & TBDC & \multicolumn{2}{c|}{AUC} & RBDC & TBDC \\
 \cline{3-4}
 \cline{7-8}
& & Micro & Macro &  &  & Micro & Macro &  & \\
 \hline 
 \hline
BMVC 2018 & Liu \etal \cite{Liu-BMVC-2018} & 84.4 & - & - & - & - & - & - & -\\
CVPR 2018 & Sultani \etal \cite{Sultani-CVPR-2018} & - & - & - & - & - & 76.5 & - & - \\
 \cline{3-4}
 \cline{7-8}
ICASSP 2018 & Lee \etal \cite{Lee-ICASSP-2018} & \multicolumn{2}{c|}{87.2} & - & - & \multicolumn{2}{c|}{76.2} & - & -\\
 \hline
WACV 2019 & Ionescu \etal \cite{Ionescu-WACV-2019} & 88.9 & - & - & - & - & - & - & - \\
ICCV 2019 & Nguyen \etal \cite{Nguyen-ICCV-2019} & 86.9 & - & - & - & - & - & - & - \\
CVPR 2019 & Ionescu \etal \cite{Ionescu-CVPR-2019} & 87.4 & 90.4 & 15.77 & 27.01 & 78.7 & 84.9 & 20.65 & 44.54 \\
 \cline{3-4}
TNNLS 2019 & Wu \etal \cite{Wu-TNNLS-2019} & \multicolumn{2}{c|}{86.6} & - & - & - & - & - & - \\
TIP 2019 & Lee \etal \cite{Lee-TIP-2019} & \multicolumn{2}{c|}{90.0} & - & - & - & - & - & - \\
 \hline
ACMMM 2020 & Yu \etal \cite{Yu-ACMMM-2020} & 89.6 & - & - & - & 74.8 & - & - & - \\
 \cline{3-4}
WACV 2020 & Ramachandra \etal \cite{Ramachandra-WACV-2020a} & \multicolumn{2}{c|}{72.0} & 35.80 & 80.90 & - & - & - & - \\
WACV 2020 & Ramachandra \etal \cite{Ramachandra-WACV-2020b} & \multicolumn{2}{c|}{87.2} & 41.20 & 78.60 & - & - & - & - \\
 \cline{7-8}
PRL 2020 & Tang \etal \cite{Tang-PRL-2020} & \multicolumn{2}{c|}{85.1} & - & - & \multicolumn{2}{c|}{73.0} & - & - \\
Access 2020 & Dong \etal \cite{Dong-Access-2020} & \multicolumn{2}{c|}{84.9} & - & - & \multicolumn{2}{c|}{73.7} & - & - \\
CVPRW 2020 & Doshi \etal \cite{Doshi-CVPRW-2020a}  & \multicolumn{2}{c|}{86.4} & - & - & \multicolumn{2}{c|}{71.6} & - & - \\
ACMMM 2020 & Sun \etal \cite{Sun-ACMMM-2020}  & \multicolumn{2}{c|}{89.6} & - & - & \multicolumn{2}{c|}{74.7} & - & -\\
ACMMM 2020 & Wang \etal \cite{Wang-ACMMM-2020} & \multicolumn{2}{c|}{87.0} & - & - & \multicolumn{2}{c|}{79.3} & - & - \\
 \hline
ICCVW 2021 & Astrid \etal \cite{Astrid-ICCVW-2021} & 84.7 & - & - & - & 73.7 & - & - & -\\
BMVC 2021 & Astrid \etal \cite{Astrid-BMVC-2021} & 87.1 & - & - & - & 75.9 & - & - & - \\
CVPR 2021 & Georgescu \etal\cite{Georgescu-CVPR-2021} &  91.5 & 92.8 & 57.00 & 58.30 & 82.4 & \textcolor{red}{90.2} & 42.80 & 83.90\\
 \hline
CVPR 2018 & Liu \etal \cite{Liu-CVPR-2018} & 85.1 & 81.7 & 19.59 & 56.01 & 72.8 & 80.6 & 17.03 & 54.23 \\
CVPR 2022 & Liu \etal \cite{Liu-CVPR-2018} + SSPCAB & \textbf{87.3} & \textbf{84.5} & \textbf{20.13} & \textbf{62.30} & \textbf{74.5} & \textbf{82.9} & \textbf{18.51} & \textbf{60.22} \\
 \hline
CVPR 2020 & Park \etal \cite{Park-CVPR-2020} & 82.8 & 86.8 & - & - & 68.3 & 79.7 & - & - \\ 
CVPR 2022 & Park \etal \cite{Park-CVPR-2020} + SSPCAB & \textbf{84.8} & \textbf{88.6} & - & - & \textbf{69.8} & \textbf{80.2} & - & - \\ 
 \hline
ICCV 2021 & Liu \etal \cite{Liu-ICCV-2021} & 89.9 & \textcolor{red}{\textbf{93.5}} & 41.05 & 86.18 & 74.2 & 83.2 & 44.41  & 83.86  \\ 
CVPR 2022 & Liu \etal \cite{Liu-ICCV-2021}  + SSPCAB &  \textbf{90.9} & 92.2 & \textbf{62.27} & \textcolor{red}{\textbf{89.28}} &  \textbf{75.5} & \textbf{83.7}  & \textcolor{red}{\textbf{45.45}} & \textcolor{red}{\textbf{84.50}} \\ 
 \hline
TPAMI 2021 & Georgescu \etal \cite{Georgescu-TPAMI-2021} &  92.3 & 90.4 & 65.05 & \textbf{66.85} &  82.7 & 89.3 & \textbf{41.34} & 78.79 \\
CVPR 2022 & Georgescu \etal\cite{Georgescu-TPAMI-2021} + SSPCAB & \textcolor{red}{\textbf{92.9}} & \textbf{91.9} & \textcolor{red}{\textbf{65.99}} & 64.91 & \textcolor{red}{\textbf{83.6}} & \textbf{89.5} &  40.55 & \textbf{83.46} \\
\hline
\end{tabular}
\vspace{-0.1cm}
\caption{Micro-averaged frame-level AUC, macro-averaged frame-level AUC, RBDC, and TBDC scores (in \%) of various state-of-the-art methods on Avenue and ShanghaiTech. Among the existing models, we select four models \cite{Georgescu-TPAMI-2021,Liu-ICCV-2021,Liu-CVPR-2018,Park-CVPR-2020} to show results before and after including SSPCAB. The best result for each before-versus-after pair is highlighted in bold. The top score for each metric is shown in red.}
\label{table:VideoAnomaly} 
\vspace{-0.3cm}
\end{table*}

\subsection{Abnormal Event Detection in Video}

\begin{figure}[!t]
\begin{center}
\centerline{\includegraphics[width=1.0\linewidth]{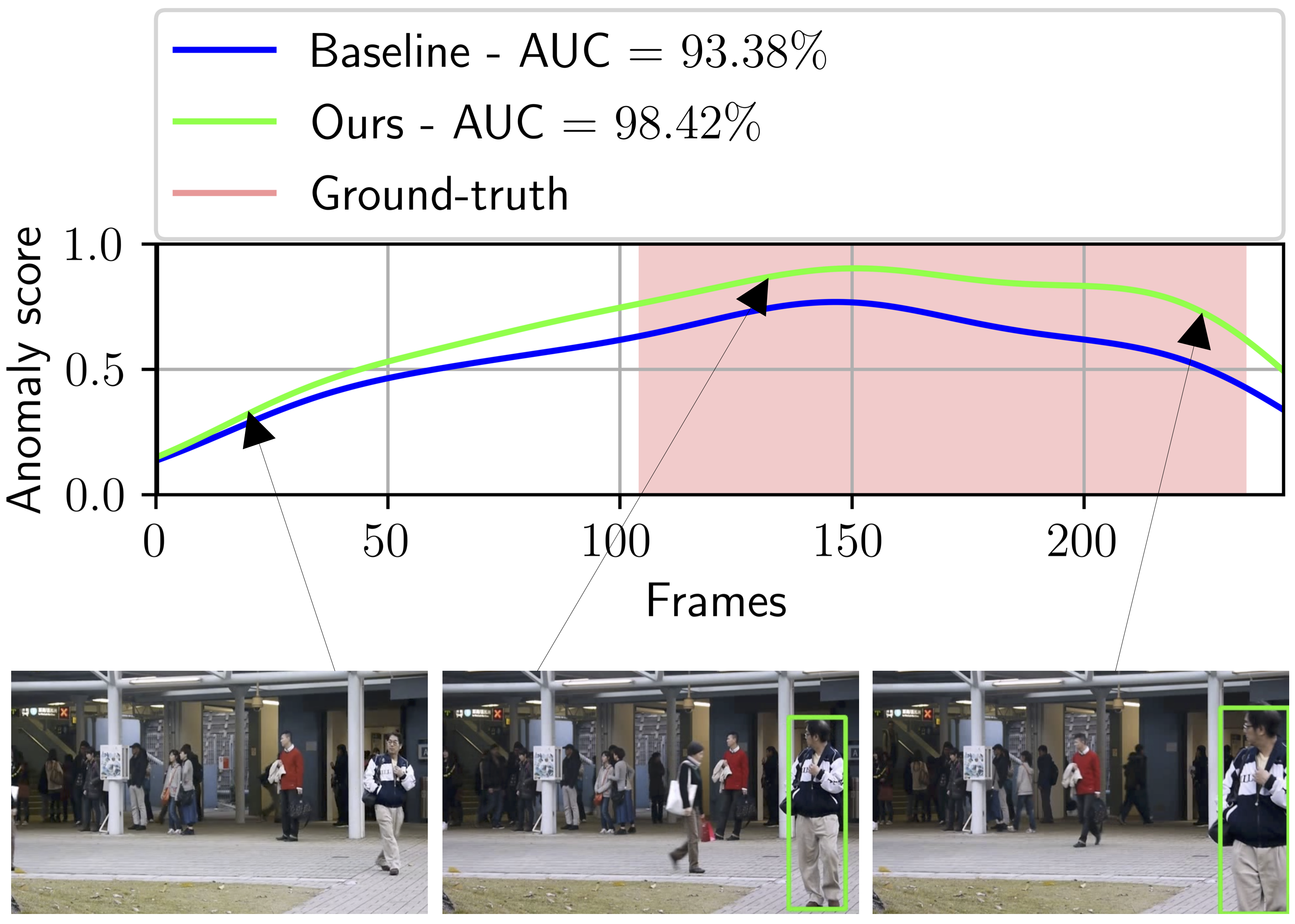}}
\vspace{-0.2cm}
\caption{Frame-level anomaly scores for Liu \etal \cite{Liu-CVPR-2018} before (baseline) and after (ours) integrating SSPCAB, for test video 18 from Avenue. Anomaly localization results correspond to the model based on SSPCAB. Best viewed in color.}
\label{fig_results_avenue}
\vspace{-0.8cm}
\end{center}
\end{figure}

\noindent
\textbf{Baselines.}
We choose four recently introduced methods \cite{Georgescu-TPAMI-2021,Liu-ICCV-2021,Liu-CVPR-2018,Park-CVPR-2020} attaining state-of-the-art performance levels in video anomaly detection, as candidates for integrating SSPCAB. We first reproduce the results using the official implementations provided by the corresponding authors \cite{Georgescu-TPAMI-2021,Liu-ICCV-2021,Liu-CVPR-2018,Park-CVPR-2020}. We refrain from making any modification to the hyperparameters of the chosen baselines. Despite using the unmodified code from the official repositories, we were not able to exactly reproduce the results of Liu \etal \cite{Liu-ICCV-2021} and Park \etal \cite{Park-CVPR-2020}, but our numbers are very close. As we add SSPCAB into the reproduced models, we consider the reproduced results as reference. We underline that, for Georgescu \etal \cite{Georgescu-TPAMI-2021}, we integrate SSPCAB into the auto-encoders, not in the binary classifiers. We report RBDC and TBDC results whenever possible, computing the scores using the implementation provided by Georgescu \etal \cite{Georgescu-TPAMI-2021}. 

\noindent
\textbf{Results.} 
We report the results on Avenue and ShanghaiTech in Table~\ref{table:VideoAnomaly}. First, we observe that the inclusion of SSPCAB in the framework of Liu \etal \cite{Liu-CVPR-2018} brings consistent improvements over all metrics on both benchmarks. Similarly, we observe consistent performance gains when integrating SSPCAB into the model of Park \etal \cite{Park-CVPR-2020}. We note that the method of Park \etal \cite{Park-CVPR-2020} does not produce anomaly localization results, preventing us from computing the RBDC and TBDC scores for their method. SSPCAB also brings consistent improvements for Liu \etal \cite{Liu-ICCV-2021}, the only exception being the macro AUC on Avenue. For this baseline \cite{Liu-ICCV-2021}, we observe a remarkable increase of $21.22\%$ in terms of the RBDC score on Avenue. Finally, we notice that SSPCAB also improves the performance of the approach proposed by Georgescu \etal \cite{Georgescu-TPAMI-2021} for almost all metrics, the exceptions being the TBDC on Avenue and RBDC on ShanghaiTech. In summary, we conclude that integrating SSPCAB is beneficial, regardless of the underlying model. Moreover, due to the integration of SSPCAB, we are able to report new state-of-the-art results on Avenue and ShanghaiTech, for several metrics.

In Figure \ref{fig_results_avenue}, we compare the frame-level anomaly scores on test video 18 from Avenue, before and after integrating SSPCAB into the method of Liu \etal \cite{Liu-CVPR-2018}. On this video, SSPCAB increases the AUC by more than $5\%$. We observe that the approach based on SSPCAB can precisely localize and detect the abnormal event (\emph{person walking in the wrong direction}). We provide more anomaly detection examples in the supplementary.


\section{Conclusion}
\label{conclusion}

In this paper, we introduced SSPCAB, a novel neural block composed of a masked convolutional layer and a channel attention module, which predicts a masked region in the convolutional receptive field. Our neural block is trained in a self-supervised manner, via a reconstruction loss of its own.
To show the benefit of using SSPCAB in anomaly detection, we integrated our block into a series of image and video anomaly detection methods \cite{Georgescu-TPAMI-2021,Liu-CVPR-2018,Li-CVPR-2021,Liu-ICCV-2021,Park-CVPR-2020,Zavrtanik-ICCV-2021}. Our empirical results indicate that SSPCAB brings performance improvements in almost all cases. The preliminary results show that both the masked convolution and the channel attention contribute to the performance gains. Furthermore, with the help of SSPCAB, we are able to obtain new state-of-the-art levels on Avenue and ShanghaiTech. We consider this as a major achievement.

In future work, we aim to extend SSPCAB by replacing the masked convolution with a masked 3D convolution. In addition, we aim to consider other application domains besides anomaly detection.

\section*{Acknowledgments}

The research leading to these results has received funding from the EEA Grants 2014-2021, under Project contract no.~EEA-RO-NO-2018-0496. This work has also been funded by the Milestone Research Programme at AAU, SecurifAI, and the Romanian Young Academy, which is funded by Stiftung Mercator and the Alexander von Humboldt Foundation for the period 2020-2022.

{\small
\bibliographystyle{ieee_fullname}
\bibliography{references}
}

\clearpage

\section{Supplementary}

\begin{table}[t!]
\centering 
\setlength\tabcolsep{4pt}
\small
\begin{tabular}{| c | c | c | c | c | c | c | c |} 
\hline
& \multicolumn{3}{|c|}{Location of SSPCAB} &  \multicolumn{2}{|c|}{AUC} & \multirow{2}{*}[0.5ex]{RBDC} & \multirow{2}{*}[0.5ex]{TBDC} \\
 \cline{2-6}
& {Early} & {Middle} & {Late} & {Micro} & {Macro} &  &  \\
 \hline
 \hline
\multirow{8}{*}[-0.5ex]{\rotatebox{90}{Plain auto-encoder}} &  &   &  & 80.0 & 83.4 & 49.98 & 51.69 \\
  \cline{2-8}
&   \checkmark  &   &  & 81.1 & 83.6 & 50.86 & 52.44\\
  \cline{2-8}
&     & \checkmark  &  & 84.2 & 85.0 & 52.73 & 54.02 \\
  \cline{2-8}
&     &   & {\color{red}\checkmark} & 85.9 & 85.6 & 53.81 & 56.33 \\
  \cline{2-8}
&   \checkmark  & \checkmark  &  & 82.7 & 83.8 & 50.54 & 52.70 \\
  \cline{2-8}
&   \checkmark  &   & \checkmark & 83.2 & 84.1 & 52.33 & 53.01 \\
  \cline{2-8}
 &    & \checkmark  & \checkmark & \textbf{86.1} & \textbf{85.7} & \textbf{54.03} & 56.07\\
  \cline{2-8}
&   \checkmark  & \checkmark  & \checkmark & 85.3 & 85.4 & 53.11 & \textbf{56.64} \\ 
\hline
\end{tabular}
\vspace{-0.2cm}
\caption{Micro-averaged frame-level AUC, macro-averaged frame-level AUC, RBDC, and TBDC scores (in \%) on Avenue, while integrating SSPCAB into an auto-encoder, at different locations. SSPCAB improves the results regardless of the integration place or the number of blocks. 
The option highlighted in red is used throughout the experiments presented in the main article. Best results are highlighted in bold.}
\label{tab_number_of_blocks}
\end{table}

\subsection{Ablation Study}

\begin{table}[t!]
\centering 
\setlength\tabcolsep{4pt}
\small
\begin{tabular}{| c | c | c | c | c |} 
\hline
Size of $M$ &  \multicolumn{2}{|c|}{AUC} & \multirow{2}{*}[0.5ex]{RBDC} & \multirow{2}{*}[0.5ex]{TBDC} \\
 \cline{2-3}
& {Micro} & {Macro} &  &  \\
 \hline
 \hline
  & 80.0 & 83.4 & 49.98 & 51.69 \\
\hline
$1\times1$ & 85.9 & 85.6 & 53.81 & 56.33 \\
\hline
$3\times3$  & 85.9 & 85.5 & 53.93 & 56.31 \\
\hline
\end{tabular}
\vspace{-0.2cm}
\caption{Micro-averaged frame-level AUC, macro-averaged frame-level AUC, RBDC, and TBDC scores (in \%) on Avenue, while varying the size of the masked kernel $M$.
}
\label{tab_size_of_M}
\end{table}

\begin{figure}[!t]
\begin{center}
\centerline{\includegraphics[width=1.0\linewidth]{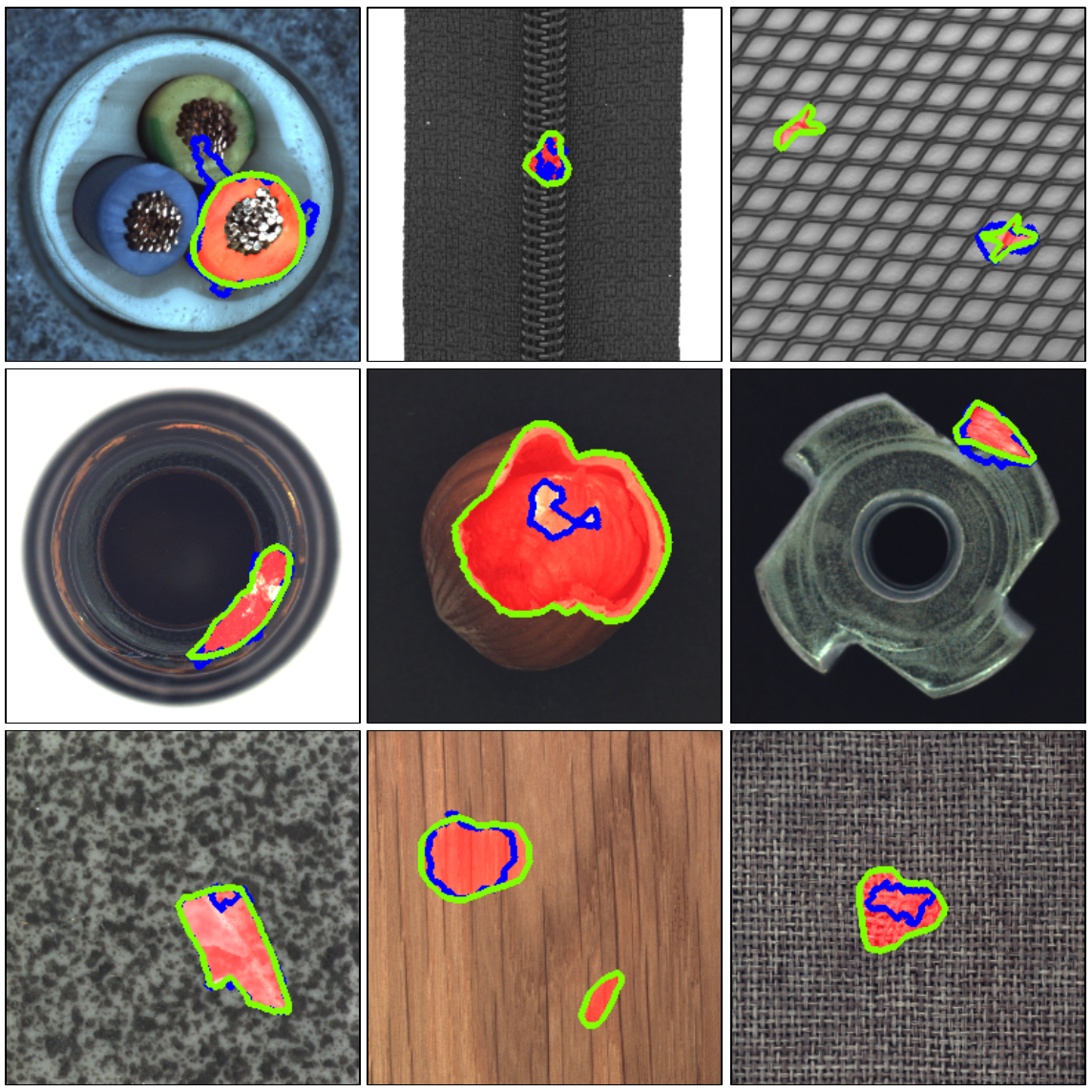}}
\caption{Additional anomaly localization examples of DRAEM \cite{Zavrtanik-ICCV-2021} (blue) versus DRAEM+SSPCAB (green) on MVTec AD. The ground-truth anomalies are marked with a red mask. Best viewed in color.}
\label{fig_results_mvtec_2}
\end{center}
\end{figure}

In the main article, we mention that we generally replace the penultimate convolutional layer with SSPCAB in underlying models \cite{Georgescu-TPAMI-2021,Liu-CVPR-2018,Li-CVPR-2021,Liu-ICCV-2021,Park-CVPR-2020,Zavrtanik-ICCV-2021}. Ideally, for optimal performance gains, the integration place and the number of SSPCAB modules should be tuned on a validation set for each framework. However, anomaly detection data sets do not have a validation set and there is no way to obtain one from the training set, as the training contains only normal examples. In this context, to fairly demonstrate the generality and utility of SSPCAB, we only used a single configuration (one block, closer to the output) across all existing frameworks. 
However, adding more modules could be beneficial. 
To test various configurations, we perform an ablation study on the number of SSPCAB modules and the places where these modules can be integrated in a plain auto-encoder. In Table~\ref{tab_number_of_blocks}, we present the corresponding experiments on the Avenue data set. \textit{We observe that SSPCAB improves the results, regardless of the place of integration or the number of blocks.} The improvements seem larger when SSPCAB is integrated closer to the output. Integrating more blocks can sometimes help.

Another hyperparameter that could be tuned is the size of the masked kernel $M$. In our experiments, we kept $M$ to a size of $1\times1$ for simplicity and speed. To study the effect of increasing the size of $M$, we have tested the size of $3\times3$ with the plain auto-encoder on Avenue. We report the corresponding results in Table~\ref{tab_size_of_M}. When comparing the results with masked kernels of $1\times1$ or $3\times3$ components, we do not observe significant differences. 

An additional aspect that can suffer multiple reconfigurations, given a validation set, is the pattern of the proposed kernel. In our experiments, we tried a simple pattern where the mask is placed in the center and the reception field is connected to the four corner sub-kernels denoted by $K_i$, $\forall i \in \{1,2,3,4\}$. We designed this pattern while trying to extrapolate the idea from middle frame prediction (which was shown to provide somewhat better results than future frame prediction) to a 2D kernel. Of course, other patterns are possible and are likely to work equally well. 

\subsection{Qualitative Anomaly Detection Results}

\noindent
{\bf Anomaly detection in images.}
In Figure~\ref{fig_results_mvtec_2}, we present additional qualitative results produced by DRAEM \cite{Zavrtanik-ICCV-2021} on the MVTec AD benchmark. The displayed examples illustrate the benefit of integrating SSPCAB, which is much better at segmenting the anomalies compared to the baseline DRAEM. We show improvements in terms of the pixel-level annotation for both objects and textures.

\begin{figure}[!t]
\begin{center}
\centerline{\includegraphics[width=1.0\linewidth]{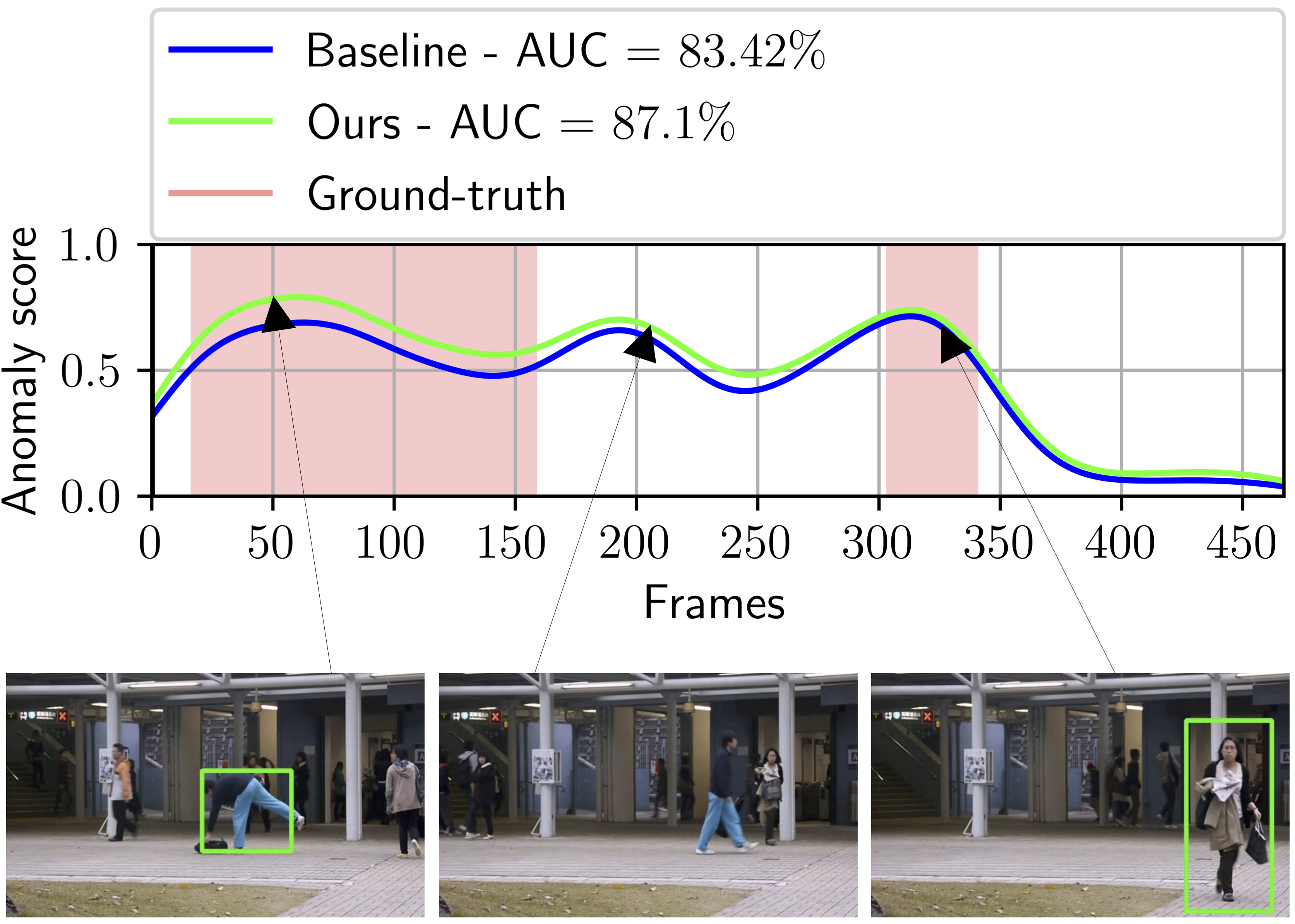}}
\caption{Frame-level anomaly scores for Liu \etal \cite{Liu-CVPR-2018} before (baseline) and after (ours) integrating SSPCAB, for test video 10 from Avenue. Anomaly localization results correspond to the model based on SSPCAB. Best viewed in color.}
\label{fig_results_avenue_10}
\end{center}
\end{figure}

\begin{figure}[!t]
\begin{center}
\centerline{\includegraphics[width=1.0\linewidth]{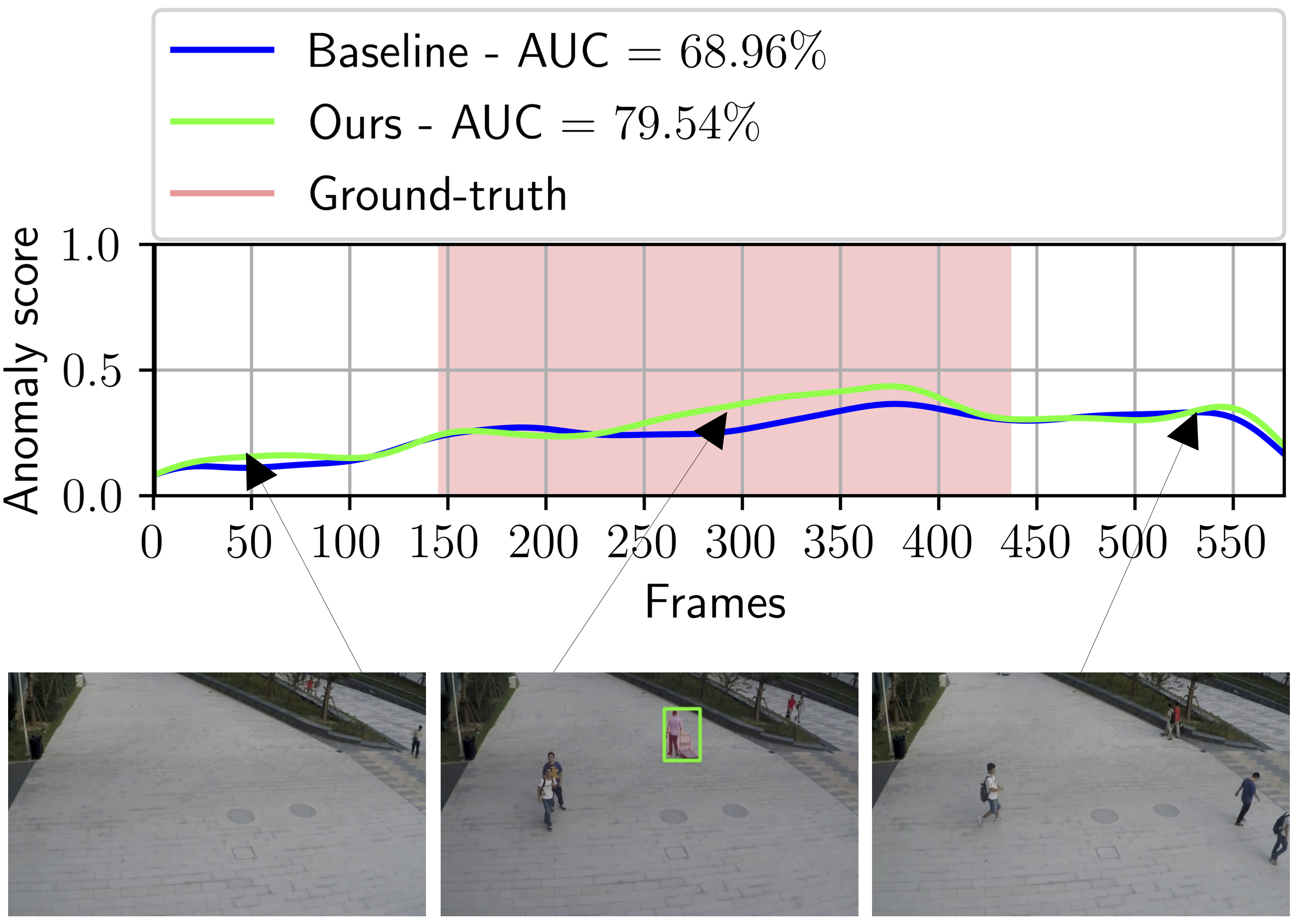}}
\caption{Frame-level anomaly scores for Georgescu \etal \cite{Georgescu-TPAMI-2021} before (baseline) and after (ours) integrating SSPCAB, for test video 01\_0054 from ShanghaiTech. Anomaly localization results correspond to the model based on SSPCAB. Best viewed in color.}
\label{fig_results_shanghai}
\end{center}
\end{figure}

\begin{figure}[!t]
\begin{center}
\centerline{\includegraphics[width=1.0\linewidth]{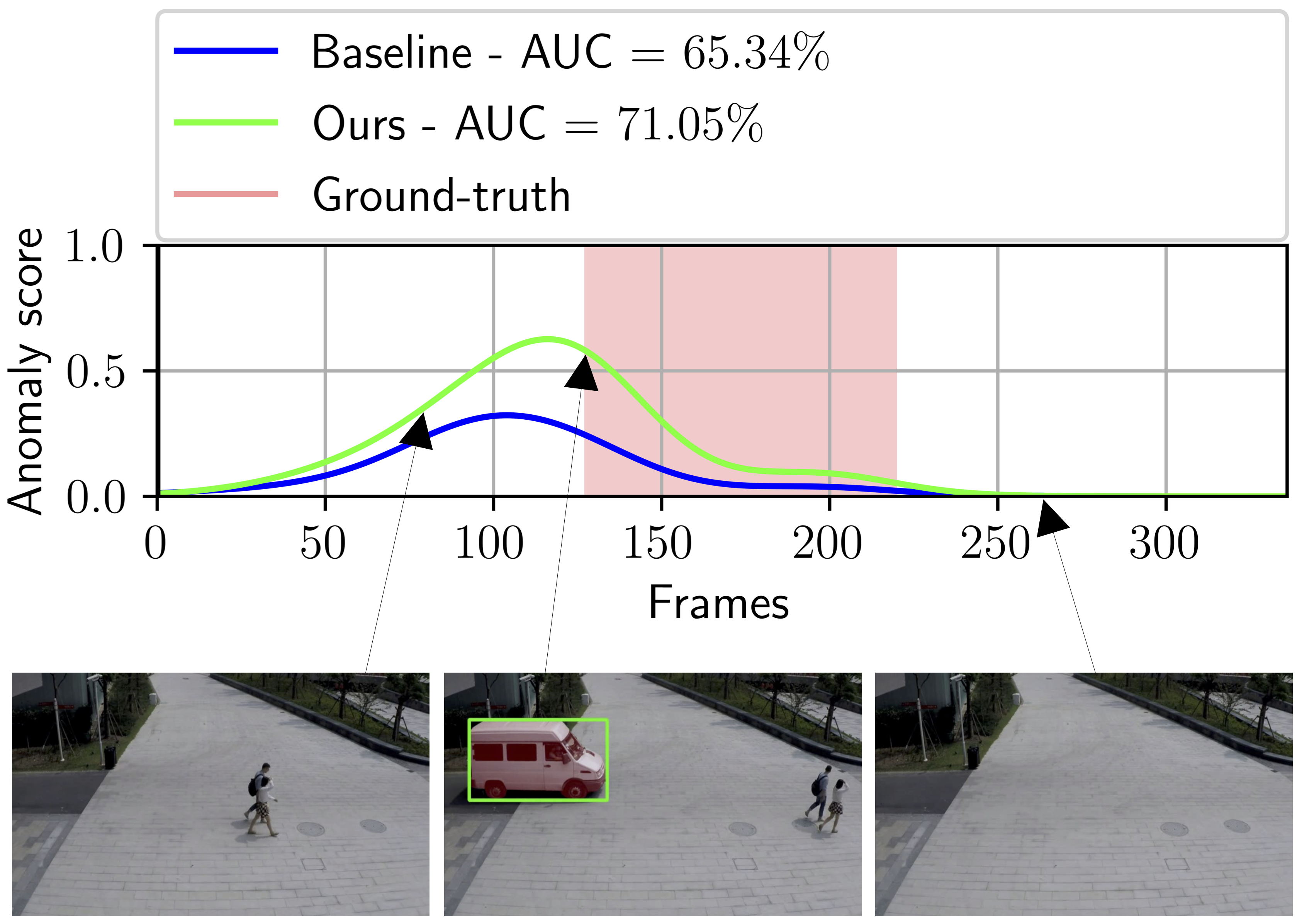}}
\caption{Frame-level anomaly scores for Georgescu \etal \cite{Georgescu-TPAMI-2021} before (baseline) and after (ours) integrating SSPCAB, for test video 01\_0130 from ShanghaiTech. Anomaly localization results correspond to the model based on SSPCAB. Best viewed in color.}
\label{fig_results_shanghai_1_0130}
\end{center}
\end{figure}

\begin{figure}[!t]
\begin{center}
\centerline{\includegraphics[width=1.0\linewidth]{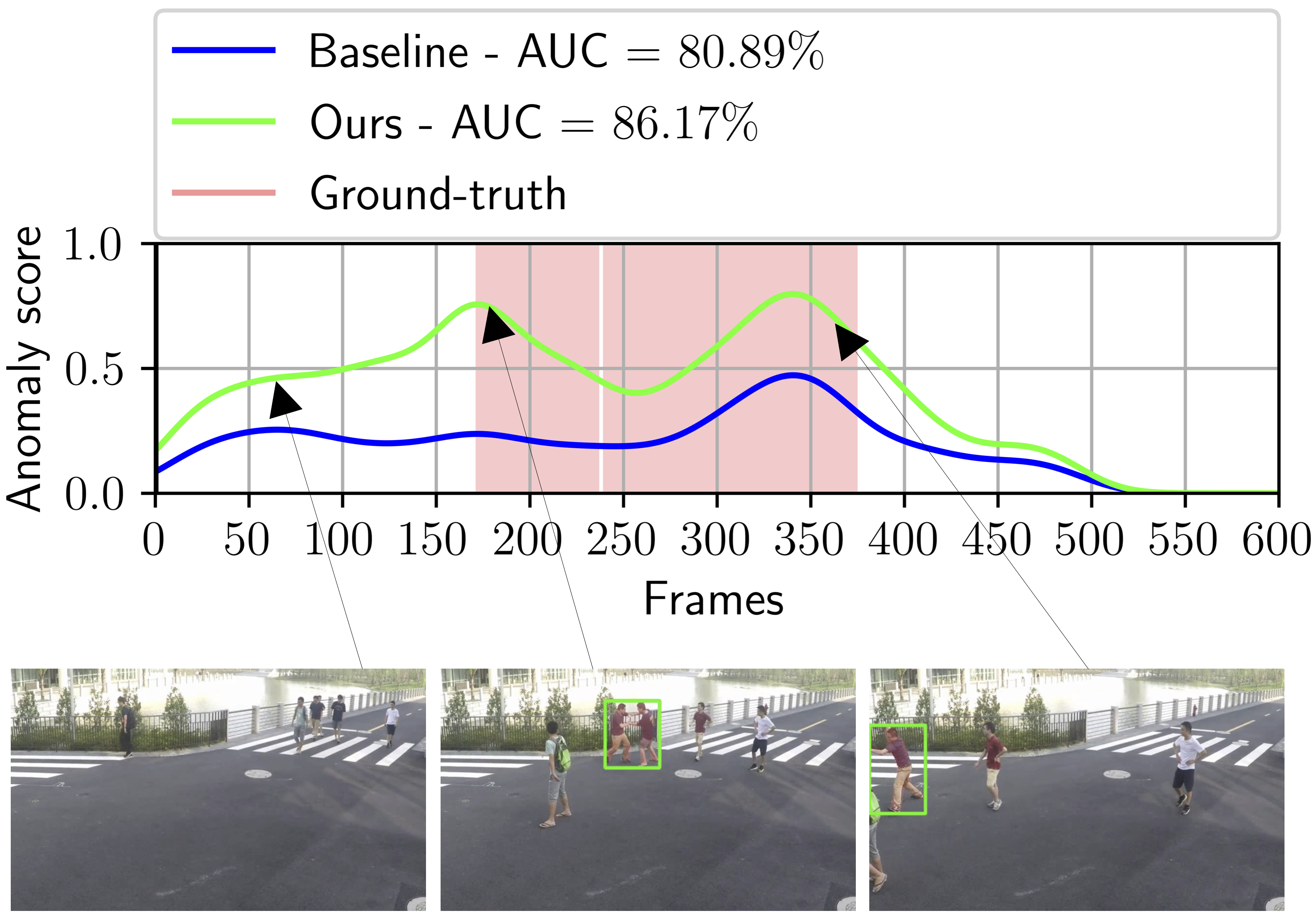}}
\caption{Frame-level anomaly scores for Georgescu \etal \cite{Georgescu-TPAMI-2021} before (baseline) and after (ours) integrating SSPCAB, for test video 07\_0047 from ShanghaiTech. Anomaly localization results correspond to the model based on SSPCAB. Best viewed in color.}
\label{fig_results_shanghai_7_0047}
\end{center}
\end{figure}

\noindent
{\bf Anomaly detection in videos.}
In Figure~\ref{fig_results_avenue_10}, we show a comparison of the frame-level anomaly scores on test video 10 from the Avenue data set, before and after integrating SSPCAB into the method of Liu \etal \cite{Liu-CVPR-2018}. On this video, SSPCAB increases the AUC by nearly $4\%$. After introducing SSPCAB, we observe higher frame-level anomaly scores for the first abnormal event. The anomaly localization results depict \emph{a person throwing a backpack} and \emph{a person walking in the wrong direction}.

In Figures \ref{fig_results_shanghai} and \ref{fig_results_shanghai_1_0130}, we illustrate similar comparisons for test videos 01\_0054 and 01\_0130 from the ShanghaiTech data set, before and after adding SSPCAB into the framework of Georgescu \etal \cite{Georgescu-TPAMI-2021}. For test video 01\_0054, SSPCAB increases the AUC by more than $10\%$. For test video  01\_0130, the baseline framework seems to detect the abnormal event too early, but SSPCAB seems capable of shifting the detection towards the correct moment. As a result, SSPCAB increases the frame-level AUC score by almost $6\%$. We observe a similar AUC improvement from SSPCAB in Figure~\ref{fig_results_shanghai_7_0047}, where we compare the frame-level anomaly scores on test video 07\_0047 from the ShanghaiTech data set. For this video, we underline that the frame-level scores are visibly more correlated to the ground-truth anomalies. Moreover, in all three ShanghaiTech videos, we observe that the approach based on SSPCAB can precisely localize and detect the abnormal events (\emph{person pulling a lever cart}, \emph{car inside pedestrian area}, \emph{people fighting}, \emph{people running}). 

\subsection{Inference Time}

\begin{table}
\centering
\noindent
\setlength\tabcolsep{2.7pt}
\begin{tabular}{|l|cc|c|}
\hline
\multirow{2}{*}{Method} & \multicolumn{2}{c|}{Time (ms)} & \multirow{2}{*}[0.5ex]{Relative (\%)} \\
\cline{2-3}
& Baseline & +SSPCAB & \\
\hline
\hline
{Liu \etal \cite{Liu-CVPR-2018}} & 2.1 & 2.4 & 14.2\\ 
\hline
{Georgescu \etal \cite{Georgescu-TPAMI-2021}} & 1.5 & 1.7 & 13.3\\
\hline
\end{tabular}
\caption{Inference times (in milliseconds) and relative time expansions (in \%) for two frameworks \cite{Liu-CVPR-2018,Georgescu-TPAMI-2021}, before and after integrating SSPCAB. The running times are measured on an Nvidia GeForce GTX 3090 GPU with 24 GB of VRAM.}
\label{tab_inference_time}
\end{table}

Regardless of the underlying framework \cite{Georgescu-TPAMI-2021,Liu-CVPR-2018,Li-CVPR-2021,Liu-ICCV-2021,Park-CVPR-2020,Zavrtanik-ICCV-2021}, we add only one instance of SSPCAB, usually replacing the penultimate convolutional layer. As such, we expect the running time to increase. To assess the amount of extra time added by SSPCAB, we present the running times before and after integrating SSPCAB into two state-of-the-art frameworks \cite{Liu-CVPR-2018,Georgescu-TPAMI-2021} in Table~\ref{tab_inference_time}. The reported times show time expansions lower than 0.3 ms for both frameworks. Hence, we consider that the accuracy gains brought by SSPCAB outweigh the marginal running time expansions observed in Table~\ref{tab_inference_time}.

\subsection{Discussion}

Although SSPCAB belongs to an existing family of anomaly detection methods, \ie reconstruction-based frameworks \cite{Fei-TMM-2020,Gong-ICCV-2019,Hasan-CVPR-2016,Li-BMVC-2020,Liu-CVPR-2018,Luo-ICCV-2017,Nguyen-ICCV-2019,Park-CVPR-2020,Ravanbakhsh-ICIP-2017,Salehi-CVPR-2021,Tang-PRL-2020,Venkataramanan-ECCV-2020}, we would like to underline that we are the first to integrate the reconstruction functionality at the block level. Unlike other reconstruction approaches, our contribution is more flexible, as it can be integrated in existing and future reconstruction methods. Moreover, SSPCAB can also be used to introduce reconstruction-based anomaly detection in other frameworks, which do not rely on reconstruction. We thus believe that our generic and effective approach will help ease future research in anomaly detection.

An important aspect that must be noted is that, due to the masked convolution, our block will not reconstruct the input exactly. Except for the degenerate case where the input is constant, this scenario should not occur in the real world, which means that the reconstruction performed by SSPCAB is not trivial. However, our foremost intuition about the usefulness of SSPCAB is different: our block provides a better reconstruction for normal convolutional features than for abnormal convolutional features. If the features representing normal versus abnormal examples are different at any layer of a neural architecture, it should result in greater differences at the final output of the architecture. This idea is also supported by the experiments presented in Table~\ref{tab_number_of_blocks}.

Further looking at the results shown in Table~\ref{tab_number_of_blocks}, we observe that SSPCAB does not bring significant gains when the block is placed near the input. We aim to further investigate this limitation in future work. Aside from this small issue, we did not observe other limitations of SSPCAB during our experiments.

\end{document}